\documentclass[letterpaper, 10 pt, conference]{ieeeconf}  

\IEEEoverridecommandlockouts                              

\usepackage{multirow}
\usepackage{svg}

\usepackage{graphicx} 
\usepackage{amsmath,amssymb,amsfonts}
\usepackage{hyperref}

\usepackage{cite}

\usepackage{subcaption}

\usepackage{booktabs}
\usepackage{color}
\newcommand*\Update{\color{black}}

\newcommand*\Done{\color{black}}

\title{\LARGE \bf
Towards Safe Maneuvering of Double-Ackermann-Steering Robots with a Soft Actor-Critic Framework
}

\author{ Kohio Deflesselle$^{1*}$, Mélodie Daniel$^{1*}$, Aly Magassouba$^{2}$, Miguel Aranda$^{3}$ and Olivier Ly$^{1}$
\thanks{*These authors equally contributed}
\thanks{$^{1}$ Univ. Bordeaux, CNRS, Bordeaux INP, LaBRI, UMR 5800 F-33400 Talence, France. $^{2}$ School of Computer Science, University of Nottingham, UK. $^{3}$ Instituto de Investigación en Ingeniería de Aragón (I3A), Universidad de Zaragoza, 50018 Zaragoza, Spain. Corresponding author: Mélodie Daniel, e-mail: \texttt{melodie.daniel@u-bordeaux.fr.}}}%

\begin{document}

\maketitle
\thispagestyle{empty}
\pagestyle{empty}


\begin{abstract}

We present a deep reinforcement learning framework based on Soft Actor-Critic (SAC) for safe and precise maneuvering of double-Ackermann-steering mobile robots (DASMRs). Unlike holonomic or simpler non-holonomic robots such as differential-drive robots, DASMRs face strong kinematic constraints that make classical planners brittle in cluttered environments. Our framework leverages the Hindsight Experience Replay (HER)  and the CrossQ overlay to encourage maneuvering efficiency while avoiding obstacles. Simulation results with a heavy four-wheel-steering rover show that the learned policy can robustly reach up to 97\% of target positions while avoiding obstacles. Our framework does not rely on handcrafted trajectories or expert demonstrations.

 \Done



\end{abstract}

\section{Introduction}

Recent advances in deep reinforcement learning (DRL) for robotics have mainly focused on navigation and locomotion tasks for robots such as omni-wheeled~\cite{Mehmood2021ITC}, quadruped~\cite{Zhang2024AS}, or biped systems~\cite{GaspardIROS2024}. By contrast, non-holonomic robots with Ackermann steering, and in particular double-Ackermann-steering mobile robots (DASMRs), remain underexplored despite their relevance in many applications such as agriculture, industrial logistics, and urban mobility~\cite{Deremetz2017ECMR}. Controlling DASMRs is significantly more complex than controlling differential-drive robots, since DASMRs cannot rotate in place and often require non-trivial maneuvers, such as temporarily moving away from the goal, to achieve correct alignment.

In the aforementioned applications, safety is as important as maneuvering performance. Robots must not only reach target positions precisely, but also avoid obstacles, adapt to uncertain dynamics, and operate robustly in cluttered environments. Classical planners such as the Timed Elastic Band (TEB)~\cite{TEBref} address safety through conservative trajectory generation, but they are highly sensitive to parameter tuning and often produce overly cautious or oscillatory behaviors~\cite{TEB_parameters_limitation}. For DASMRs, the problem is compounded by their non-holonomic constraints: limited degrees of freedom and the need to coordinate both front and rear steering make collision-free maneuvers especially challenging~\cite{Siegwart2005}.

\begin{figure} [t]
    \centering
    \captionsetup{font=scriptsize}
    \includegraphics[width=0.6\linewidth]{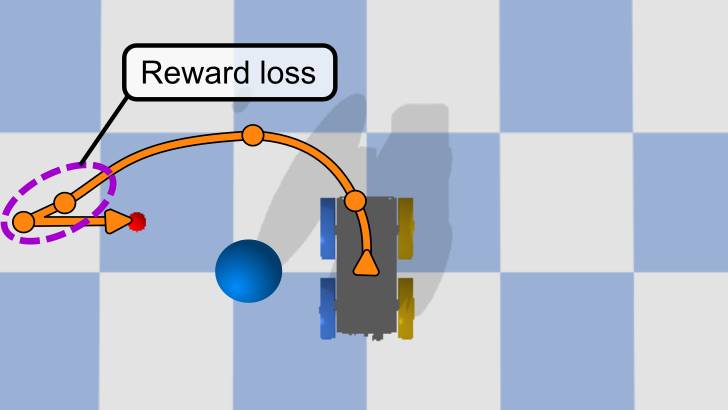}
    \caption{Maneuvering a DASMR with DRL is challenging, as it requires temporary reward loss (circled in purple), which makes classical approaches sub-optimal. The red dot indicates the desired goal, the blue sphere denotes the obstacle, and the orange curve illustrates a feasible trajectory that avoids the obstacle while reaching the goal.} \vspace{-0.6cm}
    \label{fig:eye-catch}
\end{figure}

DRL represents a promising alternative, but most existing approaches either target holonomic robots or simpler non-holonomic systems like differential-drive platforms~\cite{ZhangWTYWS25}. Reward functions based solely on distance or heading errors often penalize safe detours (cf. Fig.~\ref{fig:eye-catch}) that are required to align correctly~\cite{Lazzaroni2022APPLEPIES}. Imitation learning or curriculum learning~\cite{Honghu2022AS} are not tailored to ensure both maneuvering efficiency and obstacle-avoidance in DASMRs.


In this work, we propose an end-to-end DRL framework based on Soft Actor-Critic (SAC) combined with Hindsight Experience Replay (HER)~\cite{Andrychowicz2017HER} designed for DASMRs. Beyond enabling precise maneuvering, our framework explicitly addresses safety, with simulation validation in multiple scenarios featuring obstacles. We obtain promising results in this validation which showcase the potential of our framework.

\section{Problem Statement} \label{PS}

We consider a DASMR, controlled to make a maneuver to reach a desired 2D position $\boldsymbol{X_d} = (x_d,y_d)$. The DASMR is a non-holonomic platform. Its configuration includes $(\boldsymbol{X_c}, \theta_c)$, where $\boldsymbol{X_c}=(x_c,y_c)$ denotes the center position and $\theta_c$ the orientation of its longitudinal axis. However, only two DOF (motion along the longitudinal axis, and change of orientation) can be directly controlled, and the DASMR cannot rotate without moving forward or backward. The robot is initially positioned at the center of an 8-meter side square workspace $P$. The objective is to control the longitudinal velocity $v$ and angular velocity $\omega_c = \dot{\theta}_c$ of the robot’s center so that it reaches $\boldsymbol{X_d}$ while avoiding a fixed, known obstacle located at $\boldsymbol{X_o}$.


We restrict the workspace to a single obstacle to isolate the core difficulty of maneuvering under non-holonomic constraints while ensuring safety. This simplification is consistent with existing work such as FootstepNet~\cite{GaspardIROS2024}, which showed that such an approach can then be extended with a global planner (e.g., A*) to handle cluttered environments.


A key challenge is enabling the robot to generate feasible maneuvers without relying on prior expert knowledge or pre-defined trajectories. The problem is formulated as a Markov Decision Process~\cite{Lillicrap2015DDPG}. At each discrete time step $t$, the agent observes a state $s_t \in \mathcal{S}$, selects an action $a_t \in \mathcal{A}$ according to its policy $\pi$, and receives a reward $r_t = \mathcal{R}(s_t, a_t)$ as feedback. The policy $\pi$ is learned in simulation, with success defined by reaching the target position $\boldsymbol{X_d}$ within a distance threshold $d_{th}$ while avoiding collisions with $\boldsymbol{X_o}$.

\begin{figure}[!t]
  \centering
  \captionsetup{font=scriptsize}
  \includegraphics[width=0.7\linewidth]{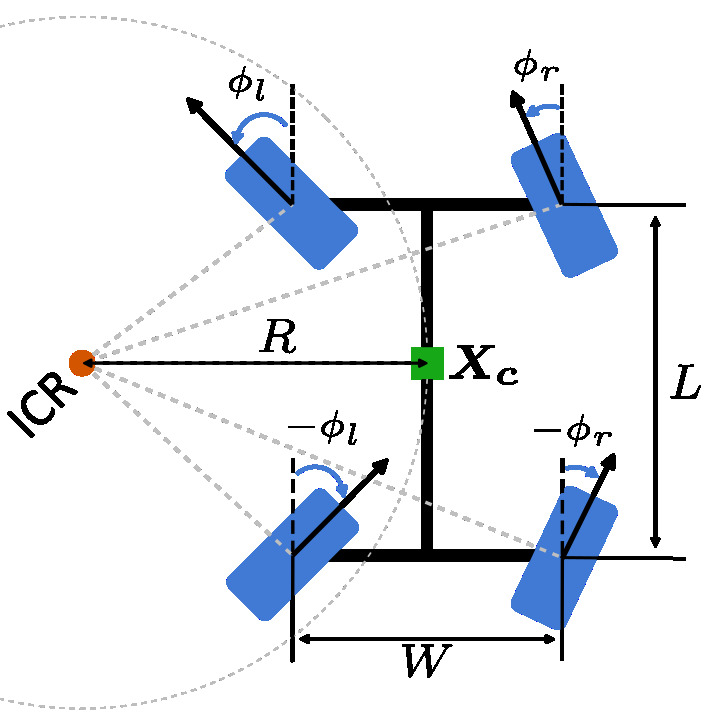}
  \caption{DASMR rotating around an instantaneous center of rotation (ICR).} \vspace{-0.5cm}
  \label{fig:ackermann-geom}
\end{figure}

\section{Method}
In our framework, a DRL agent controls $v$ and $\omega_c$. At the beginning of each training episode, the robot is initialized in the center of a square workspace containing a single fixed obstacle, and must reach $\boldsymbol{X_d}$.

\subsection{DRL Background}
In robotics, actor-critic algorithms have been successfully used in several control tasks~\cite{GaspardIROS2024, GaspardICRA2024, DanielRAL2024}. Actor-critic algorithms rely on the interaction between two dense neural networks (DNNs): an actor and a critic~\cite{DanielRAL2024}. The actor $\mu$, also called the policy network, selects an action $a_t$ based on the current state $s_t$, following $a_t = \mu(s_t)$. The critic $Q$, also called the Q-network, estimates the expected return of the state-action pair $(s_t, a_t)$ by computing the Q-value $Q^\pi (s_t, a_t)$. The critic is updated using the temporal difference learning and the Bellman equation~\cite{SuttonMIT2018}, with $Q^\pi_t (s_t, a_t) = r_t + \gamma \mathbb{E} [Q^\pi_{t+1} (s_{t+1}, a_{t+1})]$. The actor is updated by maximizing the expected Q-value.

SAC, introduced in~\cite{Haarnoja2018SAC}, jointly optimizes cumulative rewards and policy entropy to promote exploration and improve training stability. To further improve learning efficiency, the recently proposed CrossQ algorithm~\cite{Bhatt2024CrossQ} extends SAC by eliminating the need for target networks~\cite{Lillicrap2015DDPG}, through batch normalization layers. As CrossQ demonstrated promising experimental results~\cite{GaspardICRA2024}, our DRL agent uses the SAC algorithm, enhanced with the CrossQ overlay.




\subsection{State Space}
The spinning velocities of the left and right wheels are $\omega_l$ and $\omega_r$. The steering angles of the left and right wheels are $\phi_l$ and $\phi_r$, and the corresponding steering velocities $\dot{\phi}_l$ and $\dot{\phi}_r$. The robot's center linear velocity in the world frame is $\boldsymbol{V_c} = (\dot x_c, \dot y_c) = v (\cos \text{ } \theta_c, \sin \text{ } \theta_c)$. At $t$, we define the DRL agent's $\boldsymbol{s_t}$ as ($\boldsymbol{X_c}$, $\boldsymbol{X_d}$, $\theta_c$, $\omega_l$, $\omega_r$, $\phi_l$, $\phi_r$, $\dot\phi_l$, $\dot\phi_r$, $\boldsymbol{V_c}$, $\omega_c$, $\boldsymbol{X_o}$) $\in \mathbb{R}^{16}$.

\vspace{-0.cm}\subsection{Action Space}
At $t$, the DRL agent outputs an action $\boldsymbol{a_t} = (v, \omega_c) \in [-1, 1]$, representing normalized high-level velocity commands for the robot's center. These values are then scaled by the robot's respective velocity limits to obtain the actual control inputs. 

Following the double Ackermann geometry  (cf. Fig.~\ref{fig:ackermann-geom}) in a symmetric configuration (meaning rear steering angles are the inverse of front steering angles), $\omega_l$, $\omega_r$, $\phi_l$, and $\phi_r$ can be calculated from $v$ and $\omega_c$, relative to the robot's current instantaneous center of rotation (ICR) as: 
\begin{align}
    \begin{cases}
    \phi_l = tan^{-1} \frac{L/2}{R - W/2},\text{ } \phi_r = tan^{-1} \frac{L/2}{R + W/2} & \text{if $\omega \ne 0$ }\\
    \phi_l = \phi_r = 0 & \text{otherwise}
    \end{cases}
\end{align}
where $R$ is the radius to the ICR defined as $R = \frac{v}{\omega_c}$ if $\omega_c \ne 0$, $L$ is the wheelbase of the robot and $W$ is the track of the robot. Similarly, we can compute $\omega_l$ and $\omega_r$ as:
\begin{align}
    &R_l = \sqrt{(R - W / 2)^2 + (L / 2)^2} \\
    &R_r = \sqrt{(R + W / 2)^2 + (L / 2)^2}
\end{align}
\begin{align}
&\begin{cases}
    \omega_l = \frac{\omega_c}{r} R_l ,\text{ } \omega_r =\frac{\omega_c}{r} R_r & \text{if $\omega_c \ne 0$}\\
    \omega_l = \omega_r = \frac{v}{r} & \text{otherwise}
\end{cases}
\end{align}
where $r$ is the radius of the wheels, and $R_l$, $R_r$ are the left and right wheels' radii to the ICR.

\subsection{Reward Function}\label{reward_new_section}
In our formulation, we adopt a sparse reward scheme to avoid the reward loss typically induced by classical dense functions based on Euclidean distance. Such functions tend to penalize the non-trivial maneuvers and temporary reward loss that are often necessary for DASMRs to align correctly with the goal.  

To make this sparse formulation trainable, we employ HER~\cite{Andrychowicz2017HER}, which addresses the challenge of learning under sparse rewards by relabeling unsuccessful episodes. In this approach, some of the states actually reached are retrospectively treated as if they were the intended goals. This mechanism provides meaningful reward signals even when the original goal is not achieved, thereby enabling convergence.

On top of this, we introduce additional reward shaping terms to explicitly encode safety:  
\begin{equation}
    \mathcal{R}(s_t, a_t) = \begin{cases}
        -1 & \text{if $\|\boldsymbol{X_d} - \boldsymbol{X_c}\| > d_{th}$} \\
        1 & \text{if $\|\boldsymbol{X_d} - \boldsymbol{X_c}\| \leq d_{th}$} \\
        -10 & \text{if $\text{closestDistance}(\boldsymbol{X_o}) < 0.05 \text{ m}$} \\
        -100 & \text{if $\boldsymbol{X_c} \notin P$}
    \end{cases}
\end{equation}
where $\text{closestDistance}(\cdot)$ returns the closest distance between the robot and the obstacle. All norms and distances are Euclidean.

\section{Experimental Results}
\label{sec:result}

\begin{figure*}[!h]
\captionsetup{font=scriptsize}
    \centering
    \begin{subfigure}[b]{0.32\columnwidth}
        \includegraphics[width=\textwidth]{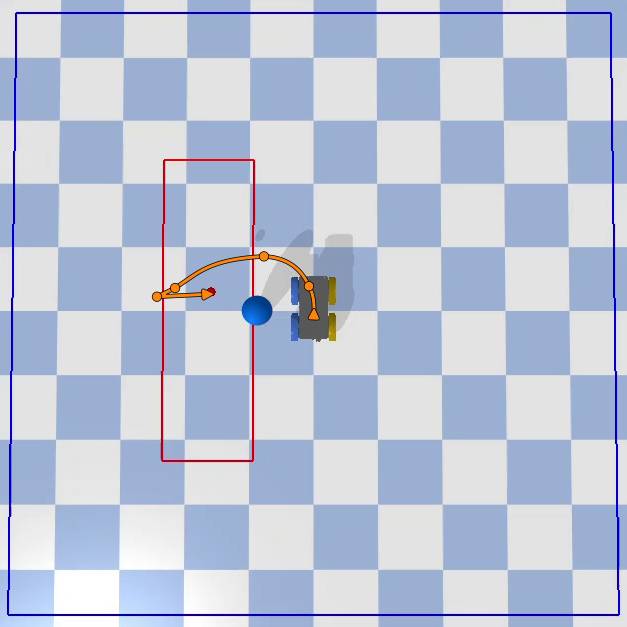}
    \end{subfigure}
    \hfill
    \begin{subfigure}[b]{0.32\columnwidth}
        \includegraphics[width=\textwidth]{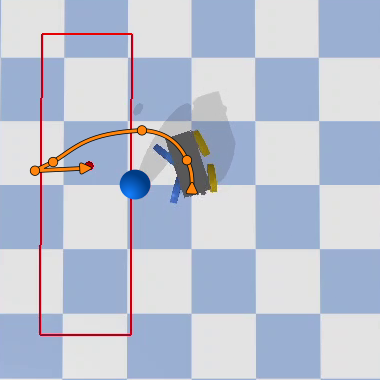}
    \end{subfigure}
    \hfill
    \begin{subfigure}[b]{0.32\columnwidth}
        \includegraphics[width=\textwidth]{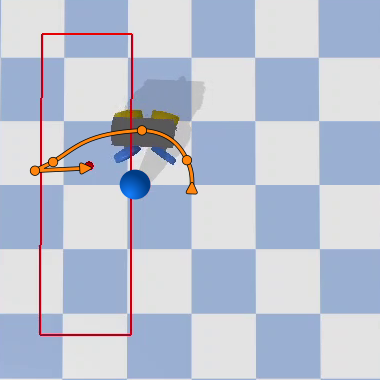}
    \end{subfigure}
    \begin{subfigure}[b]{0.32\columnwidth}
        \includegraphics[width=\textwidth]{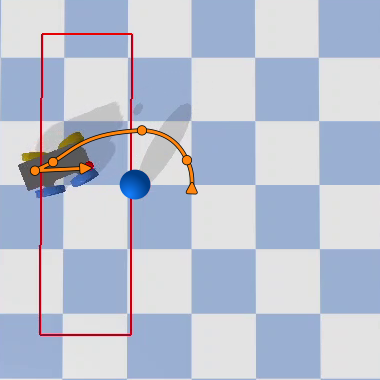}
    \end{subfigure}
    \hfill
    \begin{subfigure}[b]{0.32\columnwidth}
        \includegraphics[width=\textwidth]{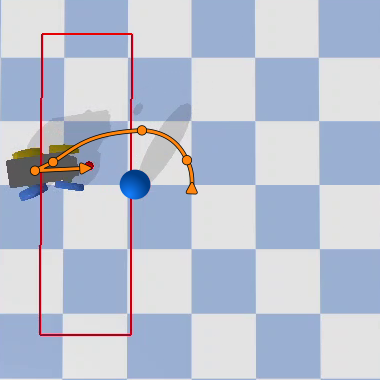}
    \end{subfigure}
    \hfill
    \begin{subfigure}[b]{0.32\columnwidth}
        \includegraphics[width=\textwidth]{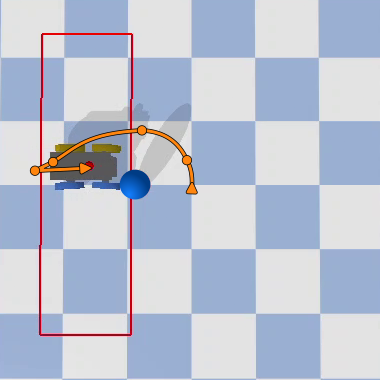}
    \end{subfigure}
    \\\vspace{-3pt}
    \dotfill
    \\\vspace{8pt}
    \begin{subfigure}[b]{0.32\columnwidth}
        \includegraphics[width=\textwidth]{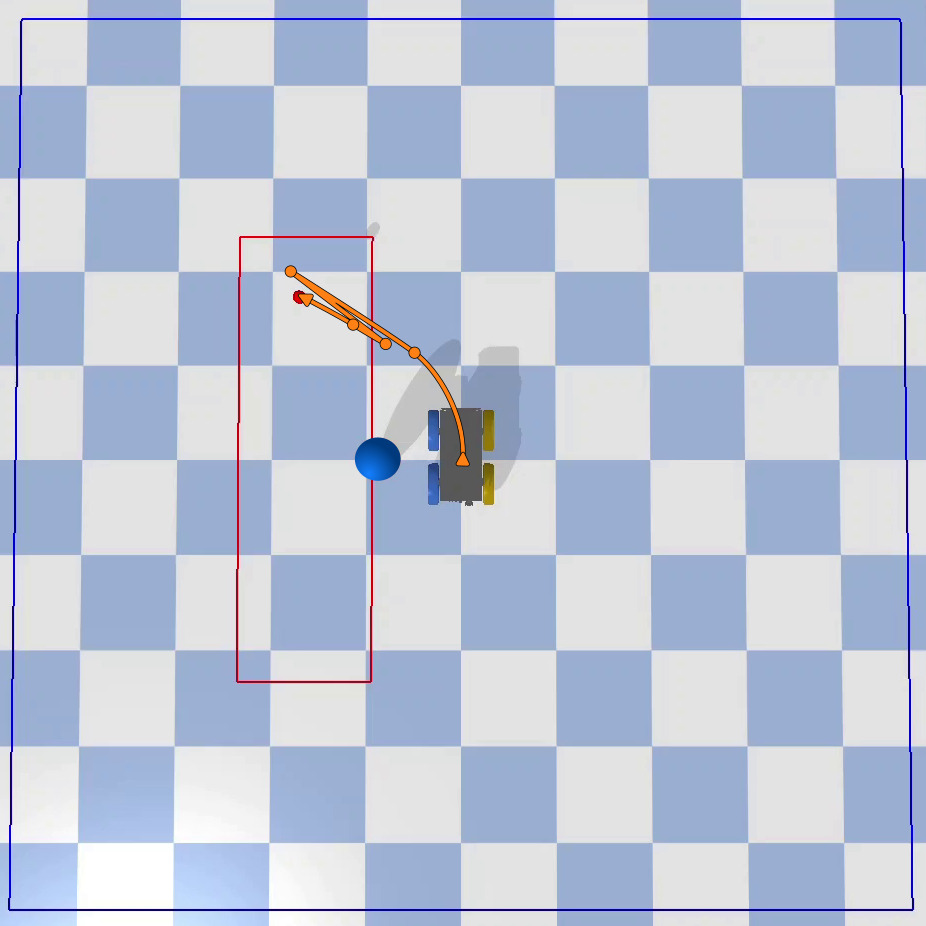}
    \end{subfigure}
    \hfill
    \begin{subfigure}[b]{0.32\columnwidth}
        \includegraphics[width=\textwidth]{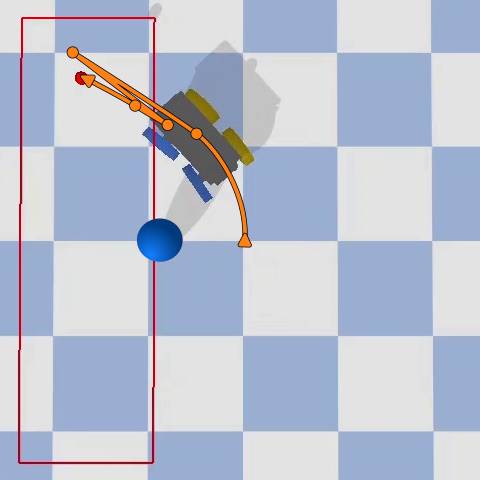}
    \end{subfigure}
    \hfill
    \begin{subfigure}[b]{0.32\columnwidth}
        \includegraphics[width=\textwidth]{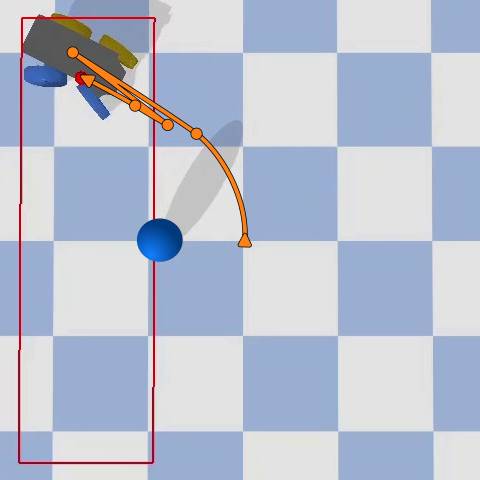}
    \end{subfigure}
    \begin{subfigure}[b]{0.32\columnwidth}
        \includegraphics[width=\textwidth]{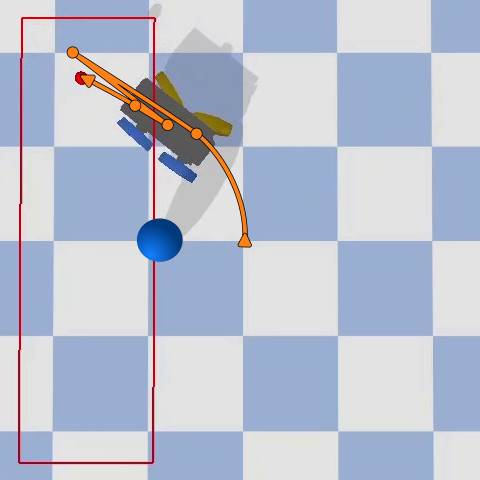}
    \end{subfigure}
    \hfill
    \begin{subfigure}[b]{0.32\columnwidth}
        \includegraphics[width=\textwidth]{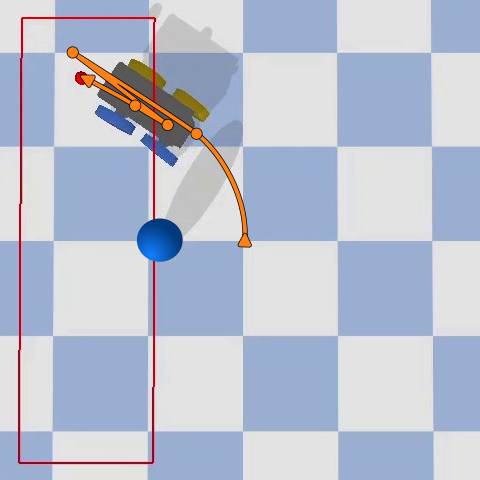}
    \end{subfigure}
    \hfill
    \begin{subfigure}[b]{0.32\columnwidth}
        \includegraphics[width=\textwidth]{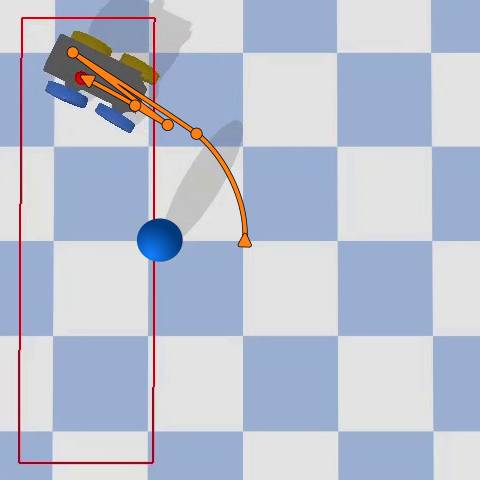}
    \end{subfigure}
    \caption{Each of the two rows shows a different maneuver executed by our agent in simulation. The red dot indicates the goal, while the blue sphere denotes the obstacle.} \vspace{-0.5cm}
    \label{fig:her-seq} 
\end{figure*}

\subsection{Environment setup}

We considered the Shadow Runner RR100 EDU rover as our mobile robot platform. The RR100 weighs approximately 100 kg and is 65 cm wide, 90 cm long, and 80 cm high. The DRL agent controlling the robot was trained using the PyBullet simulator, in an environment defined as follows: the robot is placed at coordinates $[0,0]$, and $\boldsymbol{X_o}$ is set to $(0.0,0.8)$. When the environment is reset, the robot is reset to the initial configuration, and a new $\boldsymbol{X_d}$ is sampled from a $4 \times 1.2$ m\textsuperscript{2} goal space to the left of the obstacle, with $x \in [-2, 2]$ and $y \in [0.8, 2]$. This setup ensures that most sampled targets require the robot to perform non-trivial maneuvers while avoiding the obstacle.



A training episode is considered successful when the DRL agent reaches $\boldsymbol{X_d}$ within $d_\text{th}$, regardless of orientation. If the robot fails to reach $\boldsymbol{X_d}$ within a time step limit or drives outside the boundaries of its workspace, the episode is truncated, and the environment is consequently reset. 



\begin{table}[!htb]
\captionsetup{font=scriptsize}
\caption{SAC, CrossQ, and HER parameters used on SBX during training and testing phases.}\vspace{-0.4 cm} 
\scriptsize
\label{table:hyperparams}
\begin{center}
\begin{tabular}{c|c} 
\toprule
\textbf{Parameter} & \textbf{Value} \\
\hline
Nb. layers & 2 \\
Actor Hidden size & 256 \\
Critic Hidden size & 1024 \\
$\alpha_A = \alpha_C$ & 3e-4 \\
Replay buffer size & 1,000,000 \\
Batch size $N$ & 256 \\ 
$\gamma$ & 0.99 \\ 
N. sampled goal & 16 \\
$d_\text{th}$ & 0.15 m \\
Training random seed & 9527 \\ 
Goal selection strategy & Future \\ 
Number of sampled goals & 16 \\ 
\bottomrule
\end{tabular} \vspace{-0.5 cm}
\end{center}
\end{table}


\begin{table}[!h]
\captionsetup{font=scriptsize}
\caption{\Update Benchmarking results for Seen and Unseen targets and $d_{\text{th}} = 15$ cm: with the success rate (SR) in \%, the average error (AE) (the standard deviation $\sigma$)  in m, and the SR weighted by path length (SPL). \Done} \vspace{-0.5cm}
\scriptsize
\label{table:sim-to-sim}
\scriptsize
\begin{center}
\setlength{\tabcolsep}{4pt}
\begin{tabular}{l|l|*3c}
\toprule
\multirow{2}{*}{\textbf{Approach}} & \multirow{2}{*}{$\textbf{Seed}$} & \multicolumn{3}{c}{\textbf{Metrics}}\\
\cline{3-5}
 &  & \textbf{SR $\uparrow$} & \textbf{AE($ \boldsymbol{\sigma}$) $\downarrow$} & \textbf{SPL $\uparrow$} \\                              
\hline
\multirow{2}{*} {Standard DRL~\cite{DRLStandard}}  & Seen & 29 & 0.33 (0.19) & 0.17 \\
 & Unseen & 24 & 0.43 (0.40) & 0.14 \\
\cline{1-5}
\multirow{2}{*} {Ours}  & Seen & \bf 97 & \textbf{0.17} (0.21) & \bf 0.90 \\

 & Unseen & \bf 95 & \textbf{0.15} (0.13) & \bf 0.87 \\

\bottomrule
\end{tabular} \vspace{-0.6cm}
\end{center}
\end{table}


    
    
    
    

 \subsection{Training setup}
The DRL agent was trained in a single, non-vectorized simulation environment for 300,000 time steps, where each episode was limited to 800 time steps before truncation. The agent interacted with its environment at a frequency of 40 Hz. The training progress was monitored by logging the average episode reward and success rate (SR) every 10 episodes, both computed using a sliding window of 100 episodes. The training was carried out on a desktop computer equipped with an AMD Ryzen Threadripper PRO 7985WX 64-Cores (AMD Zen 4) CPU, 128 GB of memory, along with an Nvidia RTX 4090 GPU. The DRL algorithms were implemented using the Stable Baselines JAX (SBX) library, leveraging its CrossQ implementation built upon the SAC algorithm. \Update Hyperparameters and network architecture details for SAC, CrossQ, and HER are provided in Table~\ref{table:hyperparams}, and were kept consistent in all experiments unless otherwise specified. 

\subsection{Preliminary simulation results}

To evaluate the effectiveness of our framework, we conducted simulation experiments against a classical DRL framework~\cite{DRLStandard} in two settings: (i) the same configuration as during training, and (ii) a new configuration obtained with a different random seed for goal sampling. Each framework was quantitatively evaluated with the success rate (SR), and the average distance error (AE) and the standard deviation to $\boldsymbol{X_d}$. We also evaluated the average Success weighted by (normalized inverse) Path Length (SPL)~\cite{AndersonSPL}. Higher SPL values indicate trajectories that are not only successful but also closely aligned with the shortest possible path. Achieving a high SPL is particularly challenging for DASMRs, as reaching the target position often requires complex maneuvers that deviate from the optimal path.

The results in Table~\ref{table:sim-to-sim} show a clear advantage of our framework over a standard DRL baseline~\cite{DRLStandard}. In both seen and unseen targets, our framework achieves promising success rates (97\% and 95\%), compared to less than 30\% for the baseline. This indicates that our framework generalizes well beyond the training parameters. The SPL values demonstrate that trajectories are not only successful but also significantly more efficient. Two examples of the achieved trajectories are presented in Fig.~\ref{fig:her-seq}. Importantly, the performance gap between seen and unseen settings remains small, suggesting that our framework learns robust maneuvering strategies rather than overfitting to specific goal configurations.

\section{Conclusion}\label{sec:conclusion}

This work demonstrates that our DRL framework, leveraging SAC, CrossQ, and HER, can handle the maneuvering and safety challenges of DASMRs. Preliminary results show a success rate up to 97\% and the emergence of obstacle-avoidance behaviors, confirming the potential of our framework for real-world deployment. Future work will extend the evaluation to real-world settings.



\section*{Acknowledgment}
This work was supported by the French Government under the France 2030 program through the National Research Agency (ANR) grant reference ANR-24-PEAE-0002 and the IdEx University of Bordeaux / RRI ROBSYS grant. It was also funded by the Nouvelle-Aquitaine Region through the MIRAE project.

\bibliographystyle{IEEEtran}
\bibliography{biblio_short}

\end{document}